\documentclass[letterpaper, 10 pt, conference]{ieeeconf}  
\usepackage{graphicx}
\usepackage{caption}
\usepackage{balance}
\usepackage{subcaption}
\usepackage{amsfonts}
\usepackage{siunitx}
\usepackage{booktabs}
\usepackage{cite}

\usepackage{hyperref}
\graphicspath{{./images/}}
\usepackage{amsmath}

\newcommand{\rebuttal}[1]{\textcolor{black}{#1}}
\usepackage[skip=0pt,font=small,labelfont=bf]{caption}

\IEEEoverridecommandlockouts                              

\title{\LARGE \bf
First Plan Then Evaluate: Multi-Target Planning with Post-Planning Success Evaluation Improves Learning-Based Grasping Pipelines}

\author{Martin Matak$^{1}$, Mohanraj Devendran Shanthi$^{1}$, Karl Van Wyk$^{2}$, Tucker Hermans$^{1,2}$}

\usepackage{color}
\definecolor{international_orange}{RGB}{240, 74, 0}
\definecolor{pacific_blue}{RGB}{2,108,181}

\begin{document}
\maketitle
\footnote{$^{1}$Kahlert School of Computing and the Robotics Center, University of Utah. $^{2}$NVIDIA, USA. {\tt\small martin.matak@utah.edu}}%
\begin{abstract}
    Autonomous multi-finger grasping is a fundamental capability in robotic manipulation. Optimization-based approaches show strong performance, but tend to be sensitive to initialization and are potentially time-consuming. As an alternative, the generator-evaluator-planner framework has been proposed. A generator generates grasp candidates, an evaluator ranks the proposed grasps, and a motion planner plans a trajectory to the highest-ranked grasp. If the planner doesn't find a trajectory, a new trajectory optimization is started with the next-best grasp as the target and so on. However, executing lower-ranked grasps means a lower chance of grasp success, and multiple trajectory optimizations are time-consuming. Alternatively, relaxing the threshold for motion planning accuracy allows for easier computation of a successful trajectory but implies lower accuracy in estimating grasp success likelihood.  It's a lose-lose proposition: either spend more time finding a successful trajectory or have a worse estimate of grasp success. We propose a framework that plans trajectories to a set of generated grasp targets, the evaluator estimates the grasp success likelihood at the terminal configuration of the planned trajectories, and the robot executes the trajectory most likely to succeed. 
    Our experiments show our approach improves over the traditional generator-evaluator-planner framework across different objects, generators, and motion planners, and successfully generalizes to novel environments in the real world, including different shelves and table heights. \href{https://martinmatak.github.io/fpte/}{Project Website}
    
\end{abstract}

\section{Introduction}
\begin{figure}[h]
    \centering
            \includegraphics[clip, trim=1cm 1cm 0cm 5cm, width=\columnwidth]{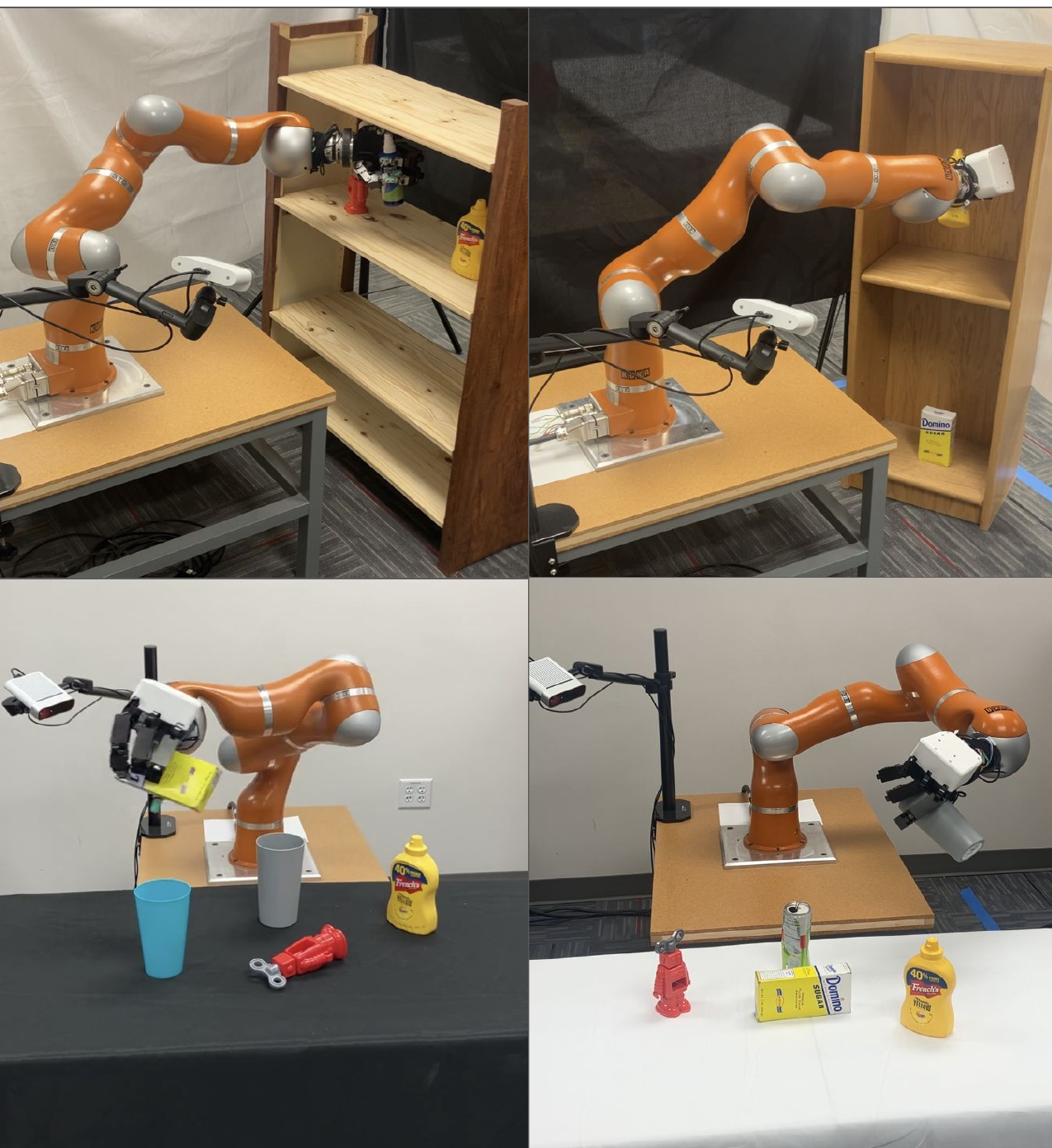}
    \caption{Grasps from different shelves and table heights are shown. The robot successfully grasps the target object while avoiding collisions with other objects and the environment. Our approach generalizes to novel environments containing multiple objects.}
    \label{fig:banner}
\end{figure}

 There is great interest in using robotic hands similar to those of humans to contend with the built environment~\cite{GAO2025102909}. Such multi-finger hands enable a robot to perform a wide range of tasks~\cite{dexterous-manipulation-rl-review}, including picking up~\cite{lu-isrr2017-grasp-inference,lu-ram2020-MultiFingeredGP,lu-ral2019-grasp-type,lu-iros2020-active-grasp,vandermerwe-icra2020-reconstruction-grasping,matak-ral23-precision-grasps,chen-implicit-shape-corl22,mandy-ngf-corl22} and manipulating objects~\cite{openai-learning-dexterous-manipulation, nvidia2022dextreme, sundaralingam-auro2019-in-grasp-optimization,bala-icra2018-gaiting}, opening doors~\cite{zhu2019dexterous}, or even playing musical instruments~\cite{robopianist2023}. In this paper, we look into multi-fingered grasping of novel objects given only a pointcloud from a single, fixed camera.

 In the area of multi-fingered grasping, optimization-based approaches that leverage a learned model to estimate the probability of grasp success are still state-of-the-art~\cite{ye2025dex1b, vandermerwe-icra2020-reconstruction-grasping, lu2023ugg, matak-ral23-precision-grasps, lu-ram2020-MultiFingeredGP} and as such tend to be sensitive to initialization and potentially computationally time-consuming. This is because optimizing over a neural network classifier is a highly nonlinear problem and multiple steps are needed to find a solution to the optimization problem. Furthermore, multiple constraints on the decision variable must exist to ensure a feasible grasp (e.g. joint limits, avoid collisions, etc.). This further complicates the process of generating grasps. To achieve faster grasp inference time, grasping as regression has been proposed~\cite{shao-ral2020-unigrasp,chen-implicit-shape-corl22} resulting in the popular generator-evaluator-planner paradigm~\cite{arsalan-icra2021-contact-graspnet, mayer-ffhnet-icra22, jens-dexdiffuser, tyler2024corl,ye2025dex1b}.
 
 The generator-evaluator-planner pipeline consists of generating a batch of grasps in the object frame, evaluating and ranking the proposed grasps, and planning a trajectory to the highest-ranked grasp. If the motion planner fails to find such a trajectory, it plans a trajectory to the next-best grasp. This leads to executing grasps that are less likely to succeed, and is computationally expensive due to running multiple trajectory optimizations in a row. Parallelization might improve computational efficiency, but executing a suboptimal grasp degrades the overall system performance.

We propose a practical and effective solution: 1) plan trajectories to all the target grasps, then 2) evaluate the grasps at the terminal configuration of the executable trajectories, and finally 3) execute the trajectory with the highest likelihood of grasp success. Since we swap the order of grasp evaluation and trajectory planning compared to the traditional approach, we name our approach \textbf{F}irst \textbf{P}lan, \textbf{T}hen \textbf{E}valuate (\textbf{FPTE}).

Furthermore, the traditional approach rejects trajectories that do not reach the target. This is because only the initial targets are evaluated on the grasp success likelihood, and proximity to the target is necessary for accurate grasp success estimation. To overcome the limitation of planning inaccuracy and evaluator sensitivity, we do not reject a collision-free trajectory if it doesn't reach the target, but evaluate the likelihood of the trajectory successfully grasping the object. This allows us to execute the trajectory with the highest likelihood of grasp success instead of discarding planned trajectories based on the somewhat arbitrary measure of distance to the proposed targets. 

Planning trajectories to all the targets is traditionally seen as inefficient. However, multiple works ~\cite{curobo_icra23, curobo_report23, dextrahg, fabrics-ral, le2024globaltensormotionplanning, 10700642} have by now addressed this issue, providing evidence for efficient computation to multiple goals. We need to plan to multiple goals only because the highest-ranked grasp is not reachable every time. Failure to successfully plan to the target grasp due to unreachability causes \textbf{20-30\% lower grasp success rates}~\cite{tyler2024corl,jens-dexdiffuser,dinesh-rss2020-deep-differentiable-planner} in the real world compared to simulation with a floating end effector or the floating object.
There are two possible explanations for this: grounding grasps in the object frame and training data for the generator.

The problem with grounding grasps in the object frame is that the generated grasps do not take robot reachability into account. On the other hand, it allows for generalization across different environments and object locations, as we show in Figure~\ref{fig:banner}. Recent RL-based approaches~\cite{dextrahg} have been successful in avoiding the object frame by learning everything in a fixed, camera frame, but it's yet to be seen how learning in a fixed frame can generalize across different object poses and environments. 
 
Training data for the generator is typically collected using a floating end-effector~\cite{turpin-eccv22-graspd, turpin-icra23-graspd, wang-icra23-dexgraspnet, tyler2024corl}. This leads to collecting grasps that might not exist in the real world due to unreachability or collisions with the environment. In an attempt to overcome this problem, using simulation, we collect a dataset of 28.9M grasp attempts by running full-robot trajectories attempting to grasp and lift the object off the table. However, in our experiments, we show that the problem of unreachable grasps persists despite the large amount of realistic data. 

As discussed before, this leads to the execution of grasps that are less likely to succeed than the top samples. On the other hand, if we relax the necessary proximity to the target pose for the motion planner, the initial grasp likelihood estimate is inaccurate as the learned evaluator is highly sensitive to the EE pose. This poses a tradeoff between the accuracy of the motion planner and the accuracy of the grasp success likelihood estimation. \textbf{FPTE} addresses this problem by evaluating grasps grounded in executable trajectories. 

Our experiments in sim and real show \textbf{FPTE} improves over the traditional generate-evaluate-plan approach across different objects, generator architectures, and motion planners. This highlights the importance of the framework independent of the specific learner or planner used. Our primary contribution lies in a subtle change to previous grasp pipelines: we evaluate grasp configurations at where the planned trajectory ends, instead of evaluating grasps proposed by the generator that serve as targets to the planner. Finally, we show successful generalization of our approach across different real world environments (Figure~\ref{fig:banner}.)

\section{Related work}
 \textbf{Grasp Planning as Optimization} Papers in this category model grasping as an optimization problem and typically use a gradient-based solver to solve it, but not always~\cite{ciocarlie-ijrr2009-eigengrasp, chen-2018-pSDF}. The metric to optimize over is either analytical~\cite{russ2018isrr, miller-icra99-force-wrench-space, ciocarlie-ijrr2009-eigengrasp, suarez-2009tro-contact-regions, suarez-2011ijrr-synthesizing-grasps, zheng-2005ijrr-force-closure-uncertainty, hang-tro2016-hierarchical-fingertip-space, siddiqui-frontiers2021-bayesian-exploration, dinesh-rss2020-deep-differentiable-planner, Maranci2024TaskOrientedGrasp, Deng2021AdaptivePlanningGrasping} or learned~\cite{lu-isrr2017-grasp-inference, lu-ral2019-grasp-type, lu-ram2020-MultiFingeredGP,lu-iros2020-active-grasp,vandermerwe-icra2020-reconstruction-grasping, matak-ral23-precision-grasps, weng-icra23-ngdf}.  While these approaches are reliable and achieve remarkable success rates, they require solving several optimizations which can be time-consuming and are sensitive to initialization. 

\textbf{Grasping as Regression}
As an alternative to optimization-based approaches, researchers have proposed approaches that generate target grasps directly without running an optimization. Redmon et al.~\cite{redmon-icra2015-real-time-grasp} reduce grasping to a computer vision problem by generating bounding boxes. Sundermeyer et al.~\cite{arsalan-icra2021-contact-graspnet} use point cloud as input and generate 6 DoF grasps, followed by execution of the most promising grasp. More recently, \cite{fuxin-icra23-generative-grasping} present an approach where they generate a grasp pose in real-time. Urain et al.~\cite{urain-se3-diffusionfields-icra23} generate trajectories to grasp objects by leveraging diffusion models. Fang et al.~\cite{fang-anygrasp-tro2023} learn a model using real-world data on a bin picking task.  Finally, Xu et al.~\cite{xu-ral21-keypoint-detection} propose adding keypoint detections to a deep network affordance segmentation pipeline.
 
\textbf{Dexterous Grasping as Regression}
Methods here extend regression-based approaches to predict full hand configurations. Our approach falls into this category. Chen et al.~\cite{chen-implicit-shape-corl22} learn a model that predicts a preshape palm pose and final grasp pose for every fingertip from a small set of human demonstrations. While~\cite{chen-implicit-shape-corl22} is specific to a four-fingered hand (although it could easily be modified), Shao et al.~\cite{shao-ral2020-unigrasp} present an approach that generates grasp contact points for any given end-effector, similarly to Wei et al.~\cite{wei2024dro}. Matak et al.~\cite{martin-icra23-workshop-ngf} present an approach to predict a grasping trajectory from a partial pointcloud, similarly to~\cite{mandy-ngf-corl22}. Mayer et al.~\cite{mayer-ffhnet-icra22} present a GAN-like approach where they take partial pointcloud as input and generate a set of grasps candidates that are ranked using a learned classifier. The basis point set they use allows for efficient encoding of the pointcloud, which we also use. Lundell et al.~\cite{lundell-icra21-multifingan} present a generative grasp sampling method that produces 6D grasps from an RGB-D image. Wei et al.~\cite{wei-ral22-dvgg} propose a generative model to generate grasps and an optimization module to refine the generated grasps. Liu et al.~\cite{dinesh-rss2020-deep-differentiable-planner} present an algorithm for training networks to grasp novel objects by using a generalized and differentiable grasp quality metric. The network is then used at runtime to predict grasps directly from a set of pointclouds.

\textbf{RL for Grasping}
Recent work in leveraging reinforcement learning for grasping~\cite{agarwal-corl2023-functional-grasping,dexrepnet,dexpoint, Zhang2024DexTOG} seems promising~\cite{dextrahg} when dealing with a single object in a known environment, though generalization to novel scenes and obstacles remains an open challenge.
In contrast, our approach generalizes to novel environments without any retraining (Figure~\ref{fig:banner}).

\textbf{Planning to Distributions}
Our work can be seen as an extension of the ideas proposed by Pavlasek et al.~\cite{pavlasek2023sets}, and Conkey et al.~\cite{adam-goal-distributions}. Instead of explicitly modeling the goal distribution, we sample from it and set the samples as the target poses for the motion planner. The grasps at the terminal configuration of the planned trajectories are then evaluated based on the likelihood of being a successful grasps, while ignoring the initial targets.

\textbf{Planning to Multiple Target Poses} 
In the area of sampling-based planners, Gammel et al.~\cite{Gammell2018InformedSampling} explicitly address the case of multiple goals, while Faroni et al.~\cite{Faroni2023OptimalTaskMotionPlanning} demonstrate that planning time is strongly sub-linear in the number of goals. Advances in accelerated computing enable a new generation of vectorized motion planning~\cite{curobo_icra23, curobo_report23,dextrahg,fabrics-ral,le2024globaltensormotionplanning,10700642}. Such planners can plan multiple trajectories in parallel without sacrificing time. Our framework yields grasp success improvement regardless of the motion planner choice. 


\section{Our Approach}
\begin{figure}[h]
    \centering
    \includegraphics[width=\columnwidth]{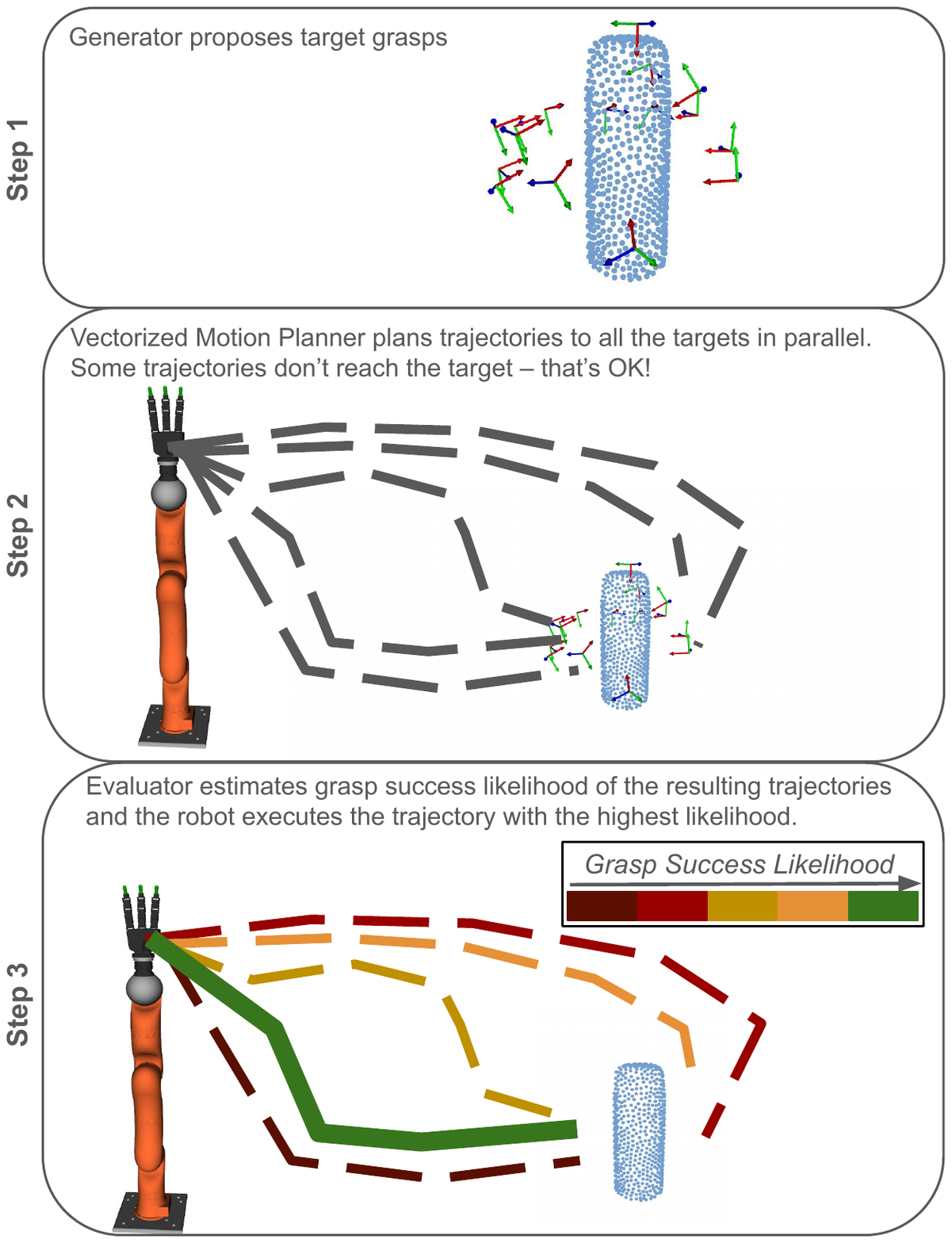}
    \caption{After the generator proposes target grasps in object frame (Step 1), we plan trajectories to all the targets and save the trajectories even if they don't reach the targets (Step 2). Then, we evaluate the likelihood of grasp success of each of the planned trajectories (Step 3) and execute the trajectory that will most likely grasp the object. Initial grasp targets are only used for the motion planner, never for grasp success likelihood estimation. Instead, we use the terminal configuration of the planned trajectories to estimate grasp success likelihood.}
    \label{fig:pipeline}
\end{figure}

FPTE consists of three stages: (1) generate grasp candidates, (2) plan trajectories to all candidates, and (3) evaluate grasp success at the terminal state of each trajectory before executing the highest-scoring trajectory (Figure~\ref{fig:pipeline})

\textbf{Step 1:} Given a partial pointcloud of the object, we pass it as input to a learned grasp generator. The generator proposes a batch of grasps in the object frame constructed from the given pointcloud. A grasp consists of a pose and hand configuration, which we formalize below.

\textbf{Step 2} A motion planner plans trajectories to all the grasps proposed by the generator while avoiding collisions with the environment and the object. Recent advances in planning trajectories to multiple targets allow for efficient computation of this step~\cite{curobo_report23, curobo_icra23, fabrics-ral}. Importantly, we ignore the distance between the resulting and the target end effector pose and do not discard any executable trajectory.

\textbf{Step 3:} A learned evaluator estimates the likelihood of grasp success at the terminal configuration of all the planned trajectories from Step 2. The grasps proposed by the generator in Step 1 are only used as targets for the motion planner in Step 2. The proposed grasps are not evaluated on the grasp success likelihood, since they may not be reached by the trajectories from step 2. Instead, grasps resulting from the planned trajectories are evaluated. Finally, the robot executes the trajectory with the highest likelihood of grasp success. 

\textbf{Object and Grasp Representation}
Let $\mathcal{O}$ be the object representation. We process the partial pointcloud using Basis Point Set (BPS)~\cite{bps} with 4096 basis points. We encode the pointcloud as the L2 norm of the distance between each of the basis points and the closest point on the pointcloud to it. This allows for a fixed-length representation $\mathcal{O} \in \mathbb R^{4096}$ that can be efficiently processed~\cite{mayer-ffhnet-icra22}.
We parametrize a grasp as $\mathcal{G} = (\textbf{T}_{OH}, \theta_p, \theta_g)$ where $\textbf{T}_{OH} \in SE(3)$ is the hand pose relative to the object, $\theta_p \in \mathbb R^n$ is a collision-free pre-grasp hand configuration, $\theta_g \in \mathbb R^n$ is an in-grasp configuration, and $n$ is the number of hand joints ($n=16$ in this work). 

\textbf{Generator} \rebuttal{The generator neural network $\mathcal{\hat{G}} = f_g(\mathcal{O})$ is trained exclusively on positive grasps.}  Given $\mathcal{O}$, $f_g$ it generates a batch of $K$ grasp samples for the motion planner. During training, we apply point-cloud perturbations to improve robustness to sensor noise, similar to~\cite{visdex,dextrahg}. 

\textbf{Evaluator} \rebuttal{The evaluator neural network $\mathcal{\hat{E}} = f_e(\mathcal{O}, \mathcal{G})$ is trained on both positive and negative samples to output the likelihood of grasp success using the cross-entropy loss.} We found the evaluator struggles to discriminate between successful and unsuccessful grasps when they are close together. To combat this, we include \textit{hard negatives}~\cite{mayer-ffhnet-icra22,arsalan-icra2021-contact-graspnet} in all minibatches by perturbing positive samples (\(\pm\) 5cm and \(\pm60^{\circ}\) per axis) and labeling them as negative. We use $\mathcal{\hat{E}}$ to evaluate the likelihood of success of the planned trajectories. Specifically, we extract $\mathcal{G}$ from the joint positions of the robot at the last timestep of a planned trajectory. 

\textbf{Grasp Planning} We use $^T$ to denote target values, and $^R$ for resulting values.
Given a set of $K$ target grasps $G_{OH}^T = \{\mathcal{G}_1^T, \mathcal{G}_2^T, ..., \mathcal{G}_K^T\}$ proposed by the generator $\mathcal{\hat{G}}$ for object $\mathcal{O}$, we transform each of the $K$ target grasp poses $\textbf{T}_{OH}^T$ from the object frame to the robot frame and obtain $K$ different $\textbf{T}_{RH}^T$. We set $G_{RH}^T=\{(\textbf{T}_{RH}, \theta_p)_1^T, (\textbf{T}_{RH}, \theta_p)_2^T, ... (\textbf{T}_{RH}, \theta_p)_K^T\}$ as targets for the motion planner. The motion planner plans $K$ trajectories from the current robot state ($\textbf{q}, \dot{\textbf{q}}$) to the targets $G_{RH}^T$ while avoiding the fixed environment  obstacles and the estimated object mesh.

Given a set of resulting trajectories $\{\tau_1, \tau_2, ..., \tau_K\}$, we extract end effector (EE) poses $\textbf{T}_{RH}^R$ and EE configurations $\theta_p^R$ at the last timestep of each trajectory $\tau_i$. We transform the EE poses $\textbf{T}_{RH}^R$ to the object frame to obtain a set of resulting EE poses $\textbf{T}_{OH}^R$  and pre-grasp configurations $\{(\textbf{T}_{OH}, \theta_p)_1^R, (\textbf{T}_{OH}, \theta_p)_2^R, ... (\textbf{T}_{OH}, \theta_p)_K^R)\}$ which we combine with $\theta_g$ values from $G_{OH}^T$ to create a set of resulting grasps $G_{OH}^R = \{\mathcal{G}_1^R, \mathcal{G}_2^R, ..., \mathcal{G}_K^R\}$. A learned evaluator $\mathcal{\hat{E}}$ estimates grasp success likelihood for each grasp in $G_{OH}^R$, and we execute $\tau_i$ that corresponds to ${G}_i^R$ with the highest likelihood. Finally, we fit a cubic spline between $\theta_p$ and $\theta_g$ to close the hand, followed by a heuristic for increasing the stiffness as in~\cite{matak-ral23-precision-grasps} before lifting the object. \rebuttal{Incidental table contact may occur while closing the hand from $\theta_p$ to $\theta_g$.}

\textbf{Grasp Generation}
Our dataset contains objects with associated grasps, rendered depth images, and grasp success labels.  The grasp data is stored as tuples $\{(\mathcal{G}^j, \mathcal{O}^j, y^j)\}$ where $\mathcal{O}$ is pointcloud rendered from the single, fixed depth camera, and $y$ is a binary label. To synthesize a sample, we spawn an object from the BigBird dataset~\cite{singh-icra2014-bigbird} at a random orientation around the z axis at a random position on the table. We render the depth camera and save the pointcloud. We generate $K$ target grasp poses around the object and pre-grasp hand configurations using a heuristic~\cite{matak-ral23-precision-grasps}. We then use a vectorized motion planner~\cite{fabrics-ral} to move the arm to the target poses and configurations while avoiding collisions with the object and table. Each environment in the simulator executes one of the $K$ trajectories. We opted for this solution over a floating end effector because this approach generates realistic grasps for the whole embodiment.

Given the resulting collision-free hand configuration $\theta_p$, we plan for a hand configuration $\theta_g$ such that the fingertips of the hand are in contact with the object surface as in~\cite{matak-ral23-precision-grasps}. We solve the problem in parallel for all the environments in the simulator using a vectorized motion planner~\cite{fabrics-ral}. Once we compute the configuration $\theta_g$, we fit a cubic spline between $\theta_p$ and $\theta_g$ and execute the trajectories, followed by a heuristic for increasing the stiffness as in~\cite{matak-ral23-precision-grasps} before lifting.

During data collection, we use fabrics~\cite{fabrics-ral} as the motion planner~\cite{dextrahg, nvidia2022dextreme}. We plan to a batch of $K$ different grasps $\mathcal{G}$. During data collection, these $K$ target grasp poses are generated using a heuristic described previously. We execute all $K$ trajectories, one per environment, and label the data. 

Our grasp generation pipeline is motivated by Matak et al.~\cite{matak-ral23-precision-grasps}, but heavily modified to (1) leverage parallelized simulation while executing full-robot trajectories, and (2) plan precision grasps during data generation instead of at runtime as in~\cite{matak-ral23-precision-grasps}. 
We generated 28.9M grasp attempts in simulation using full-robot trajectories planned to heuristic grasp proposals. Successful and failed grasps were retained for training, yielding approximately 180k successful grasps.

\section{Experiments}
We conduct two sets of experiments to help us understand the system: in simulation and in the real world. In simulation, we focus on understanding the efficiency of our system in a known environment. In the real world, we focus on generalization across novel objects and novel environments. All of our experiments are performed on a KUKA LBR4 (7 DoF) with a four-fingered Allegro hand (16 DoF).

\begin{figure}[h]
    \centering
    \includegraphics[width=0.6\columnwidth]{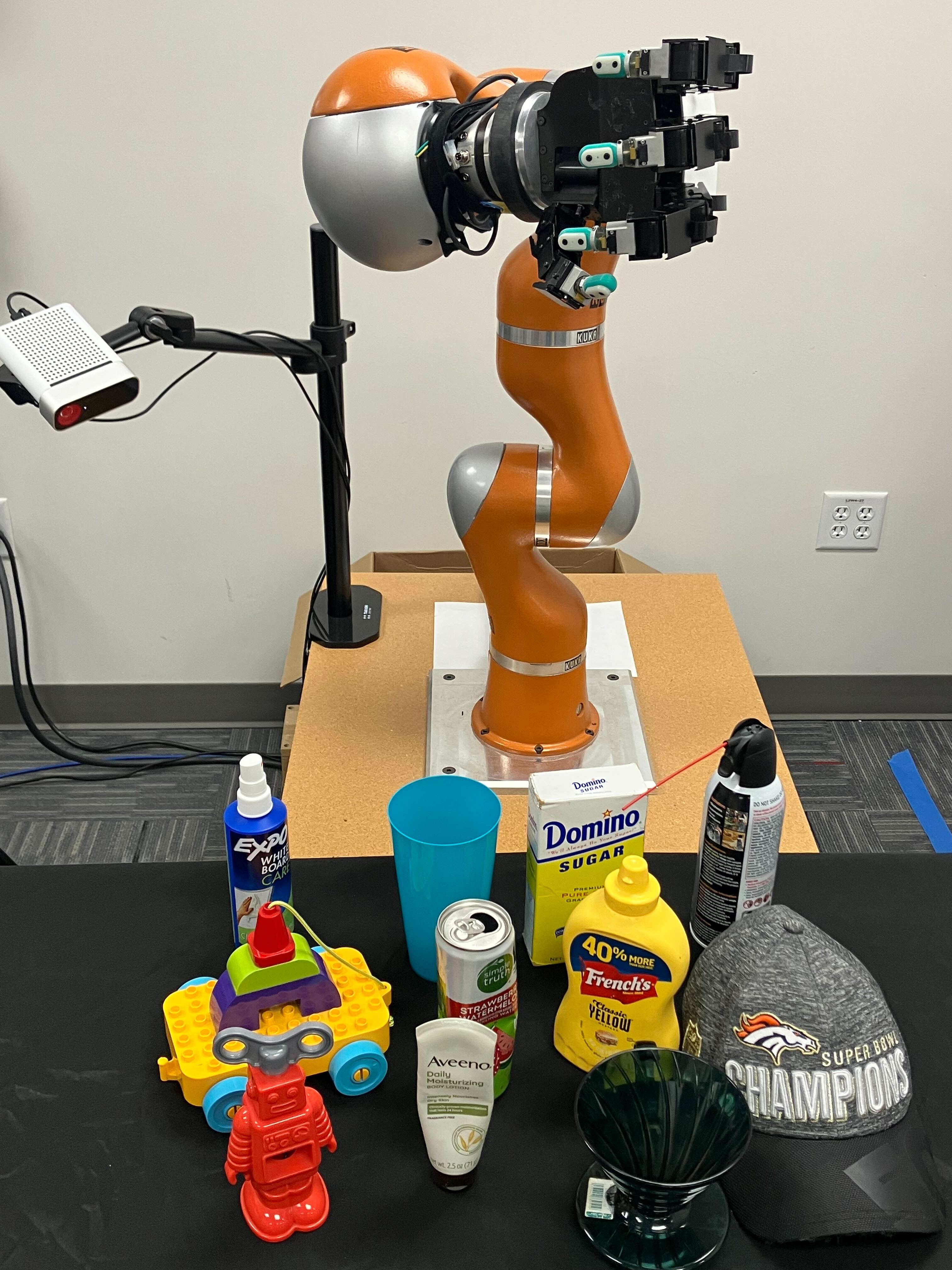}
    \caption{Our platform and the objects used to assess our system. The objects vary in shape, size, mass, rigidity, and opaqueness.}
    \label{fig:rw-objects-used}
\end{figure}

\subsection{Simulation: understanding the system}
\begin{figure}[h]
    \centering
    \includegraphics[width=\columnwidth, trim={1cm 0 0 0},clip]{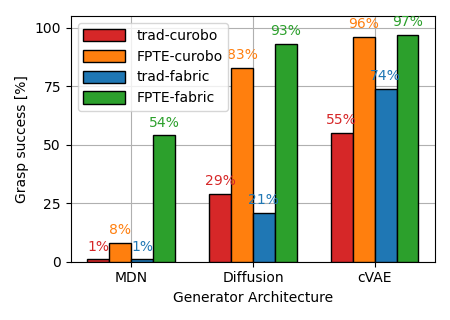}
    \caption{Comparison of grasp success rates for different generators, motion planners, and approaches. Our approach, \textbf{F}irst \textbf{P}lan \textbf{T}hen \textbf{E}valuate, consistently achieves a higher success rate compared to the ``trad'' method, irrespective of the motion planner choice or the generator architecture.}
    \label{fig:success-rates}
\end{figure}

To understand the impact of our approach on the overall robot performance, we compare our approach to the traditional generate-evaluate-plan approach. In the traditional approach, after the generator proposes grasp targets and the evaluator ranks the grasp targets, the motion planner plans a trajectory to the highest-ranked grasp. If the planner \textit{succeeds}, the robot executes the trajectory. If not, the motion planner plans to the next-best ranked grasp. We use default position (0.5 cm) and rotation (14 degrees per axis) threshold values from Curobo~\cite{curobo_report23} to determine if the planned trajectory succeeds. In Fig.~\ref{fig:success-rates}, we refer to this method as ``trad''.

\begin{figure}[h]
    \centering
    \includegraphics[width=\columnwidth]{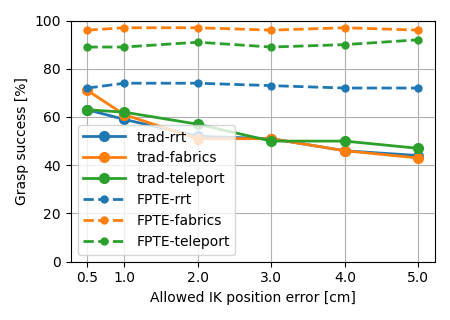}
    \caption{FPTE yields higher success rates regardless of the IK solution accuracy or the motion planner choice. Only the traditional method degrades in performance as the IK error increases. In case IK error is zero, the traditional approach and FPTE result in the same grasp, and thus success rate (not visualized).}
    \label{fig:ik-ablation-success-rates}
\end{figure}

Instead of planning trajectories one by one, we use a vectorized motion planner to plan trajectories to all grasp targets simultaneously. This does not slow down planning since all computations are done in parallel on a GPU~\cite{curobo_report23, fabrics-ral}. After planning, we rank all the trajectories based on their grasp likelihood. This is in contrast to the ``trad'' method that ranks grasp targets proposed by the generator; hence, \textbf{FPTE} -- \textbf{F}irst \textbf{P}lan, \textbf{T}hen \textbf{E}valuate. Importantly, we rank all the resulting trajectories, not only \textit{successful} trajectories. This allows the robot to execute the trajectory that is most likely to succeed in grasping the object, even if distinct from the initially generated grasp.

We conduct experiments across two different vectorized motion planners (Curobo~\cite{curobo_report23} and Fabrics~\cite{fabrics-ral}), three different generator architectures (cVAe~\cite{mayer-ffhnet-icra22}, Diffusion~\cite{tyler2024corl, jens-dexdiffuser}, and MDN~\cite{mohan-ral-pickandplace}), and 20 different object shapes from the training dataset across 100 poses on the table. We set the batch size to 128 for Fabrics~\cite{fabrics-ral} and 64 for cuRobo~\cite{curobo_report23} since it requires more GPU memory. We keep the object shape encoding method (BPS) and the learned evaluator model the same throughout all the experiments.

We use Isaac Sim as our physics simulator. A single object is placed in isolation on the table at one of the 100 poses, and a fixed depth sensor is used to observe the partial pointcloud. The observed pointcloud is used as the input for the generator, and the loaded mesh of the object is used for collision modeling for the motion planner.

\textbf{Results} are presented in Figure~\ref{fig:ik-ablation-success-rates}. We observe that FPTE-fabrics achieves higher success rates than FPTE-curobo. We hypothesize this is because fabrics was used during data collection, thus limiting the gap between data collection and deployment. We observe FPTE-rrt has lower success rates than other FPTE approaches. This is because of coarse collision checking, which allowed more experiments but resulted in more collisions during execution.

Importantly, our approach, \textbf{F}irst \textbf{P}lan \textbf{T}hen \textbf{E}valuate, consistently achieves a higher success rate irrespective of the motion planner choice or the generator architecture. To understand why this is the case, we ablate the importance of distance to the target during motion planning on grasp success while keeping the generator (cVAE), the evaluator, and the rotation threshold fixed.

We compare three different motion planners: RRT~\cite{rrt}, fabrics~\cite{fabrics-ral}, and \textit{teleportation} directly to the IK solution. We use a custom IK solver for RRT and \textit{teleportation}. We visualize the results in Figure~\ref{fig:ik-ablation-success-rates}. Analyzing results for the \textit{teleportation} method, we observe FPTE yields higher grasp success rates compared to the traditional method regardless of the motion planner, as the robot is directly teleported to the collision-free IK solution. Results using RRT and fabrics confirm that finding.

\begin{figure}
   \centering
    \includegraphics[trim={0 0 0 0mm}, clip, width=\linewidth]{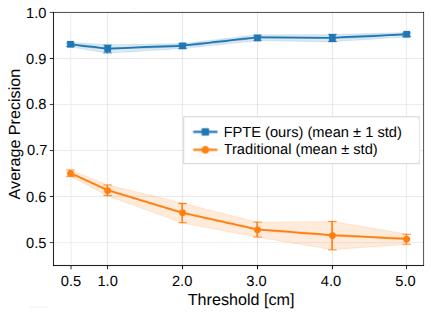}
    \caption{Evaluator average performance (AP) across different IK position thresholds. Using the traditional pipeline, the evaluator performs worse as the IK threshold increases. This is not the case when our approach is used.}
    \label{fig:ik_ablation_evaluator_performance}
\end{figure}

Furthermore, we analyze the evaluator's precision as the allowed IK position increases and observe degradation in average precision as the IK position error increases only when the traditional method is used, as visualized in Figure~\ref{fig:ik_ablation_evaluator_performance}. The observed evaluator sensitivity to IK error in the traditional approach, but not in FPTE, explains why FPTE outperforms the traditional approach.

\begin{figure}[h]
    \centering
    \includegraphics[width=\columnwidth]{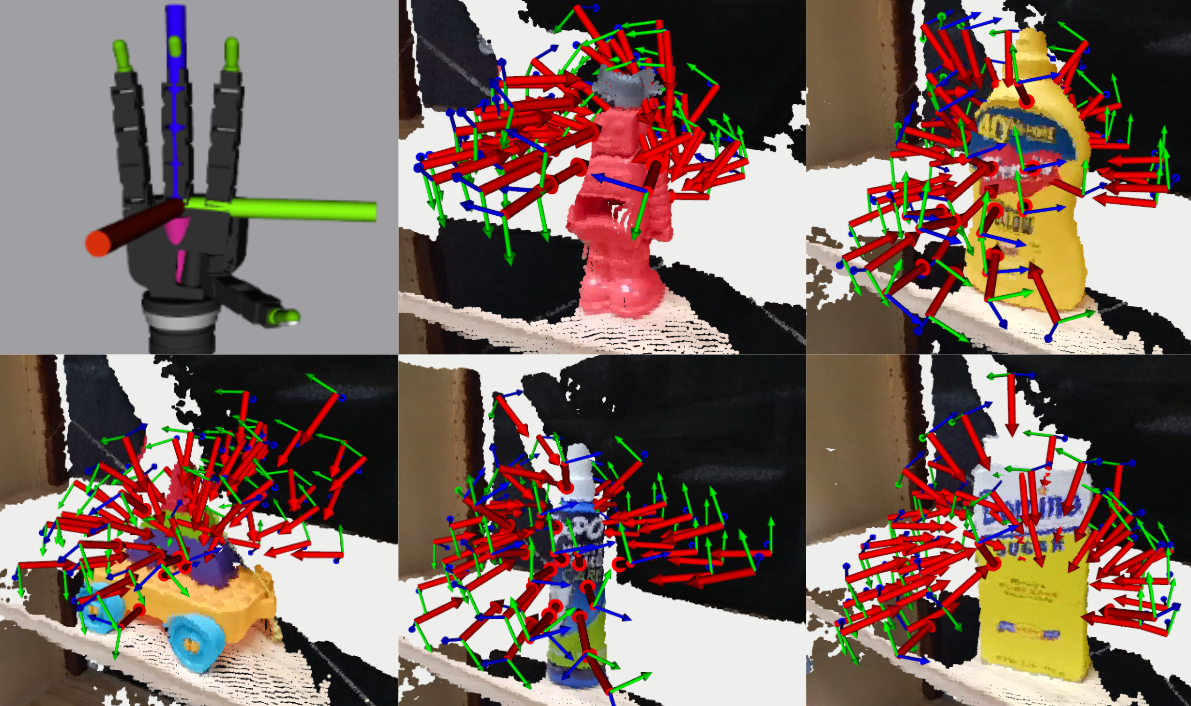}
    \caption{Our generator successfully captures a diverse grasp distribution and generalizes to novel objects in the real world.}
    \label{fig:rw-grasp-distribution}
\end{figure}

We observe unreachable grasps across different generators and motion planners, despite using physically realistic data for training and the generated grasp targets being distributed around the object, as shown in Figure~\ref{fig:rw-grasp-distribution}. The unreachability of the proposed grasps stems from grounding grasps in the object frame. However, grounding grasps in the object frame allows for generalization across different object poses and environments, which we analyze next. 

\subsection{Real World Generalization}
We evaluate our approach across 11 novel objects shown in Figure~\ref{fig:rw-objects-used}. We place an object on the table, segment it, reconstruct its mesh, plan a grasp trajectory, and execute it.

\begin{figure}[h]
    \centering
\includegraphics[width=\columnwidth,clip,trim={0cm, 6cm, 0cm, 0cm}]{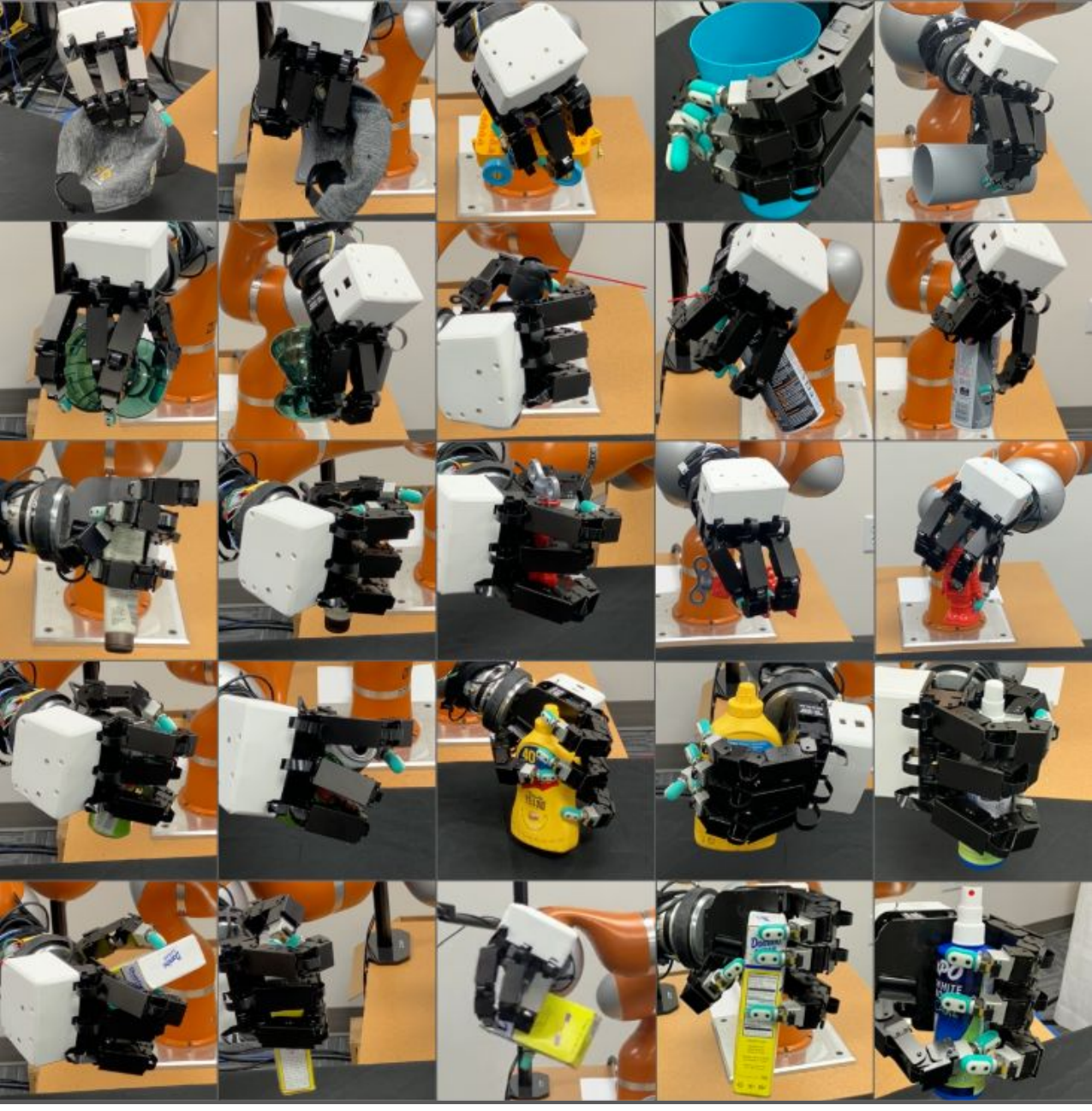}
    \caption{Some examples of resulting grasps.}
    \label{fig:rw-grasps}
\end{figure}

We use Grounded SAM~\cite{ren2024grounded}, on the RGB image from the depth observation with the prompt \textit{object that can be picked up with one hand}, followed by matching the RGB segmentation with depth values. We use BRRP~\cite{hebie2025brrp} to reconstruct the object mesh from the partial view and model it as an obstacle for the motion planner. We use the best-performing duo from the simulator: cVAE~\cite{mayer-ffhnet-icra22} as the generator, and fabrics~\cite{fabrics-ral} as a motion planner with batch size 512. We place every object across five different locations on the table. If the baseline cannot find a successful trajectory after three attempts, we count it as a failed attempt. Our approach, \textbf{FPTE}, achieves an 80\% success rate compared to 22\% with the traditional approach. It is worth noting that the baseline didn't find a trajectory for 23 attempts. When counting only executed trajectories, the success rate of the baseline was 38\%. The difference in success rate comes from 1) the baseline not finding a trajectory (23 out of 55 times), and 2) the executed grasps are less likely to succeed as evaluated by the learned evaluator. On average, the predicted likelihood of success of executed grasps using our approach was 90\% compared to 26\% when using the baseline. That is because the baseline ignores trajectories that would lead to a good grasp if the distance to the target pose is above the goal threshold. Some of the resulting grasps are shown in Figure~\ref{fig:rw-grasps}.

\begin{figure}[ht]
\centering
\begin{subfigure}[b]{\columnwidth}
  \includegraphics[width=1\linewidth]{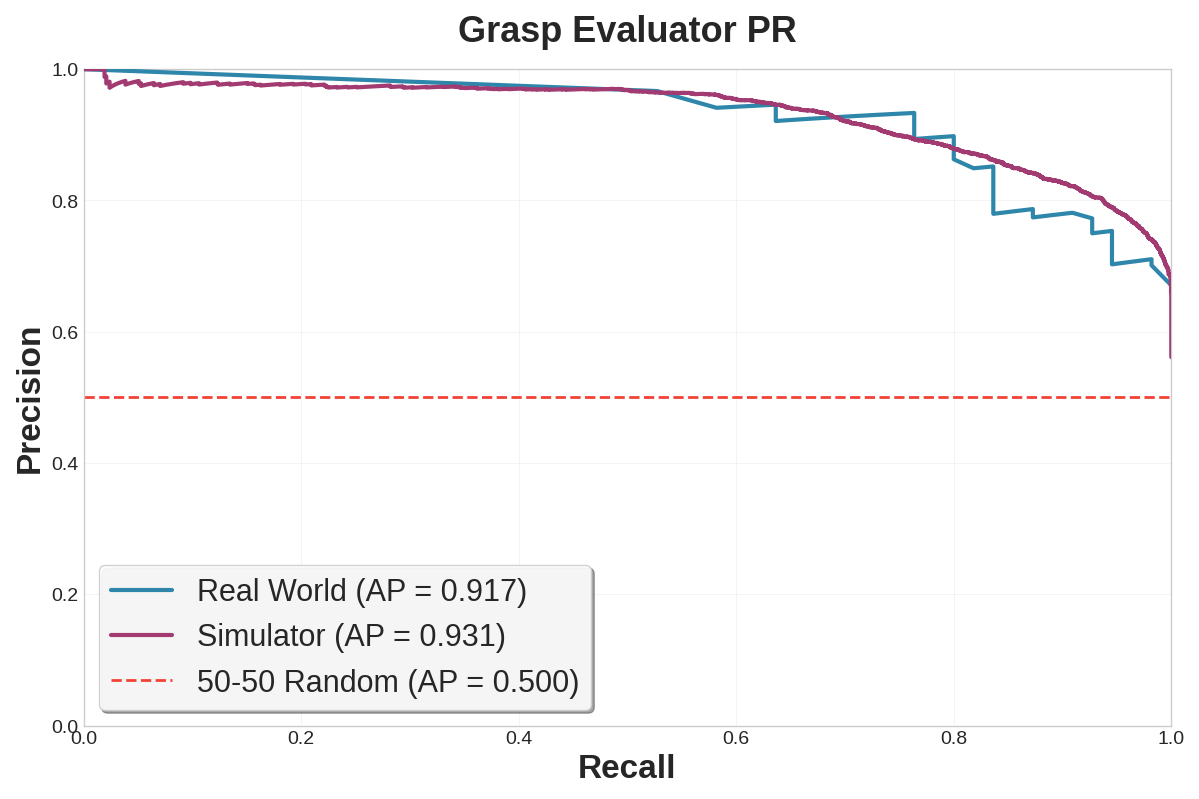}
\end{subfigure}
\begin{subfigure}[b]{\columnwidth}
  \includegraphics[width=1\linewidth]{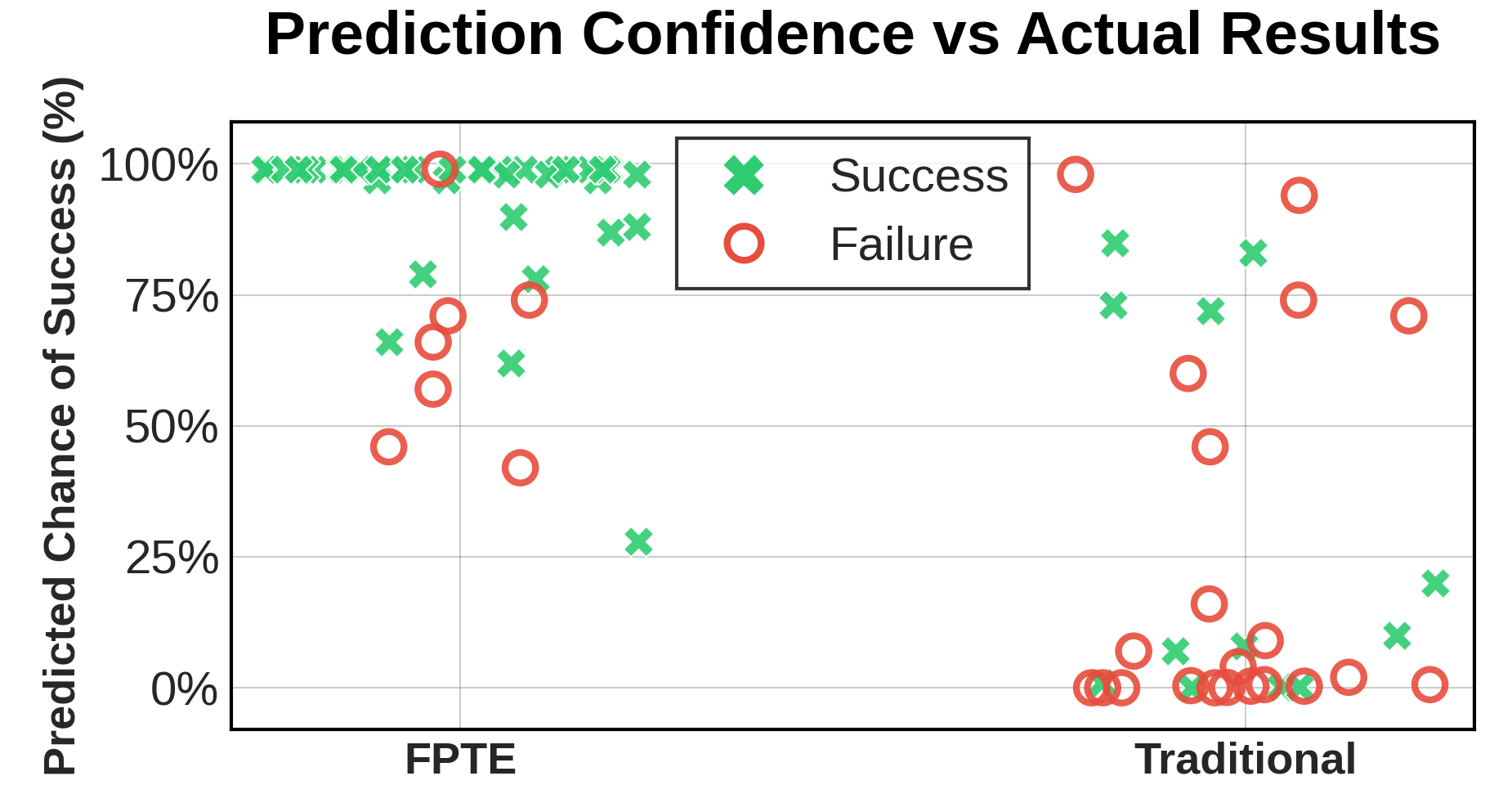}
\end{subfigure}
\caption{%
\textbf{Top} Precision–recall of the evaluator in simulation and real world. \textbf{Bottom} Predicted success vs. outcomes. FPTE produces higher-confidence successful grasps, suggesting thresholding could further improve performance.
}
\label{fig:pr-evaluator}
\end{figure}

Finally, to assess the soundness of our evaluator, we plot Figure~\ref{fig:pr-evaluator}. We report (top) the precision-recall (PR) curve of our evaluator used to estimate the likelihood of grasp success. In plotting the PR curve, we filter out samples where the robot hit the object while moving to the pre-grasp configuration.  We further visualize (bottom) predictions of the evaluator in the real world for the executed grasps.

Figure~\ref{fig:pr-evaluator} shows us that the evaluator 1) performs well in the real world, despite being trained on simulated data only, and 2) grasp success rates can be further increased by executing grasps only if the evaluator's prediction is above a high threshold. For example, we could execute grasps only if the likelihood of success is above 90\% and call for human intervention otherwise. To construct the plots, in our experiments we didn't filter out any grasps based on the estimated probability of success. 

\textbf{Generalization to Novel Environments} Object-centric grasp planning enables us to grasp the target object even if there are multiple objects on the table. Furthermore, it enables our approach to grasp objects in novel environments, such as different table heights and shelves. Without measuring quantitative results, we show different environments and grasps in Figure 1. We model all the detected objects and the environment (hand-coded) as collisions for the motion planner. Due to inaccuracies in the learned object shape reconstruction~\cite{hebie2025brrp}, the robot occasionally made unintended contact with the object while approaching the target grasp. We observed 8 such contacts across 25 executed trajectories, and 2 of these contacts led to grasp failure. Additionally, the robot's fingertips contacted the table during hand closure in 4 trials; however, none of these contacts led to grasp failure. Note that all of the learned models have been trained solely on data collected by grasping a single object in isolation from a fixed table height in the simulator. The real-world objects used are not part of the training data, and clutter has never been simulated during training. 

Note that all of the learned models have been trained solely on data collected by grasping a single object in isolation from a fixed table height in the simulator. The real-world objects used are not part of the training data, and a cluttered environment has never been simulated during training.

\section{Conclusion \& Limitations}
In this work, we propose a novel framework that improves learning-based grasping pipelines. By evaluating the likelihood of success at the terminal configuration of planned trajectories, instead of evaluating proposed grasp targets that may not be reachable, we improve grasp success rates.

Limitations include limited training-object diversity, and segmentation inaccuracy that leads to incorrect collision modeling \rebuttal{and thus potential contact with objects modeled as collisions in a cluttered scene.} The object shape diversity can be addressed using a larger dataset~\cite{ye2025dex1b}, potentially combined with active learning~\cite{lu-iros2020-active-grasp}. On the perception side, the system runs in open loop based on a single RGBD reading. 

In our experiments, we show that our approach leads to improved grasp success rates over the traditional approach. We show our improvements persist across different objects, generator architectures, and motion planners. Furthermore, we show the ability of our method to generalize in the real world across novel objects and scenes that may contain multiple objects, despite training only by grasping objects in isolation off a fixed table height in simulation. 





\section*{ACKNOWLEDGMENT}
This work was supported by DARPA under grant HR0011-24-9-0423.
\balance
\bibliographystyle{IEEEtran}
{
\footnotesize
\bibliography{references}
}

\end{document}